\def\year{2020}\relax
\documentclass[letterpaper]{article} 
\usepackage{aaai20}  
\usepackage{times}  
\usepackage{helvet} 
\usepackage{courier}  
\usepackage[hyphens]{url}  
\usepackage{graphicx} 
\usepackage{graphicx}  
\usepackage{amsmath}
\usepackage{amssymb}
\usepackage{pifont}
\usepackage{multirow}
\usepackage{threeparttable}
\usepackage{booktabs}
\usepackage{multirow}
 \usepackage{verbatim}
 \newcommand{\cmark}{\ding{51}}
\newcommand{\xmark}{\ding{55}}
\newcommand{\citet}[1]{\citeauthor{#1} \shortcite{#1}}
\newcommand{\citep}{\cite} 
\newcommand{\citealp}[1]{(\citeauthor{#1}, \citeyear{#1})}

\title{Facial Action Unit Intensity Estimation via Semantic Correspondence Learning with Dynamic Graph Convolution}
\author{  \Large \textbf{Yingruo Fan, Jacqueline C.K. Lam, Victor O.K. Li}\\ 
Department of Electrical and Electronic Engineering,\\
The University of Hong Kong, Hong Kong\\
 \{yingruo,jcklam,vli\}@eee.hku.hk
}

\begin{document}

\maketitle

\begin{abstract}
The intensity estimation of facial action units (AUs) is challenging due to subtle changes in the person's facial appearance. Previous approaches mainly rely on probabilistic models or predefined rules for modeling co-occurrence relationships among AUs, leading to limited generalization. In contrast, we present a new learning framework that automatically learns the latent relationships of AUs via establishing semantic correspondences between feature maps. In the heatmap regression-based network, feature maps preserve rich semantic information associated with AU intensities and locations. Moreover, the AU co-occurring pattern can be reflected by activating a set of feature channels, where each channel encodes a specific visual pattern of AU. This motivates us to model the correlation among feature channels, which implicitly represents the co-occurrence relationship of AU intensity levels. Specifically, we introduce a semantic correspondence convolution (SCC) module to dynamically compute the correspondences from deep and low resolution feature maps, and thus enhancing the discriminability of features. The experimental results demonstrate the effectiveness and the superior performance of our method on two benchmark datasets.
\end{abstract}

\section{Introduction}
Facial action unit intensity estimation aims to estimate intensity levels of facial muscle actions, named Action Units (AUs). Progress in this area can facilitate the applications of chatbot~\cite{lin2019caire,lin2020xpersona}, empathic design~\citealp{lin2019moel}, facial expression recognition (FER)~\citealp{fan2018multi}, etc. In real-world scenarios, human facial expressions can be deconstructed using the varied combinations of AUs and their intensities. Therefore, estimating intensity levels of AUs is an important step for further interpreting facial expressions. To provide a comprehensive encoding method, the facial action coding system (FACS)~\citealp{eckman1978facial} defines rules for scoring AU intensity on a six-point ordinal scale.  However, large-scale acquisition of facial AU intensity data is often difficult and time consuming, since trained specialists are required to annotate the data. Furthermore, the facial appearance changes are subtle in terms of AU intensity, and individuals may have different levels of expressiveness due to their own physical characteristics. The six-level intensities data of AUs are also highly imbalanced as the highest level occurs rarely. Thus, compared to FER~\cite{fan2018video,fan2020facial} and AU recognition tasks~\citealp{li2019semantic}, discerning different intensities of AUs remains a far more challenging task.

\begin{figure}
\centering
\includegraphics[width=.95\columnwidth]{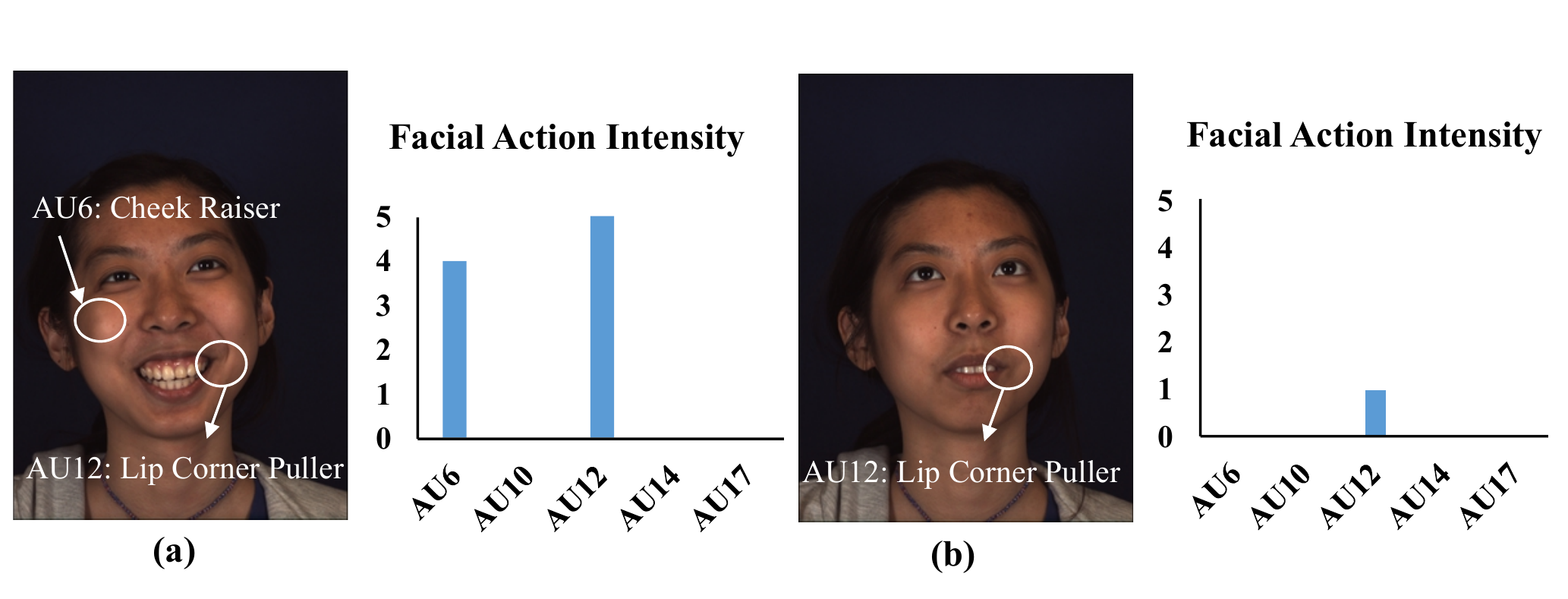}
\caption{\label{fig:AU} Co-occurrencs of different AU intensity levels. (a) High intensity AU12 
activates the occurrence of AU6; (b) Low intensity AU12 occurs independently.}
\end{figure}

The co-occurrences of different AU intensity levels may influence the overall estimation results. The facial appearance changes caused by a certain AU are also affected by the intensities of other AUs. As an example in Figure~\ref{fig:AU}, AU6 (Cheek Raiser) and AU12 (Lip Corner Puller) are typically activated together in the case of the happy expression. More specifically, a high intensity of AU12 can increase the probability of AU6 occuring and vice versa. By contrast, a low intensity of AU12 might present independently without activating other AUs. 

To infer the co-occurrences of AUs, prior studies undertaken by Walecki et al.~\shortcite{walecki2016copula} and Wang et al.~\shortcite{wang2018facial} are typically based on probabilistic graphical models that directly learn the AU dependencies. Intuitively, the feature level information should contain more comprehensive descriptions than the final outputs.
Hence, we formulate the problem into a heatmap regression-based framework, where the feature maps preserve rich semantic information associated with AU intensities and locations. We infer that through activating various feature channels simultaneously, the framework would produce the final AU co-occurrence pattern accordingly. 
This conjecture naturally leads to the idea of using the semantic correspondence between feature channels for discovering the latent co-occurrence relationships of AU intensities. The key novelty of our method is that such relationships are automatically captured via dynamically computing the correspondences from feature maps layer by layer, rather than relying on AU probabilities as in previous approaches. 

In this study, one motivation lies in the phenomenon that the specific AU with a certain intensity can cause visible facial appearance changes. This can be reflected by a heatmap, which is generally utilized to visualize the spatial distribution of a response. Heatmap regression has proven to be notably effective~\citealp{sanchez2018joint}, and thus our framework inherits the advantage of the heatmap-based scheme for AU intensity estimation. In our framework, the intensity level of each AU is learned from the response at its corresponding location on the heatmap. For example, if there is a larger response, we can infer that this location may belong to an AU with higher intensity. 

The other motivation comes from graph convolutional neural networks (GCNNs)~\citealp{scarselli2008graph}, which are proposed to pass and aggregate information in the graph structure. Due to the advantages in extracting the discriminative feature representations, GCNNs have been widely used in image classification~\citealp{wang2018zero}, semantic segmentation~\citealp{liang2017interpretable}, relational reasoning~\citealp{battaglia2016interaction}, etc. More importantly, GCNN describes the intrinsic relationship between various vertex nodes of the graph by learning an adjacency matrix, thus providing a potential way for us to explore the relationships among multiple feature maps. In our method, we introduce a simple yet effective semantic correspondence convolution module, dubbed SCC, which automatically learns the semantic relationships among feature maps. We summarize the key contributions of our work as follows:

\romannumeral1 ) We propose to leverage the semantic correspondence for modeling the implicit co-occurrence relations of AU intensity levels in a heatmap regression framework, where the feature maps encode rich semantic descriptions and spatial distributions of AUs.

\romannumeral2 ) We introduce a semantic correspondence convolution module to dynamically compute the correspondences among feature maps layer by layer. To our knowledge, this is the first work that brings the advantages of dynamic graph convolutions to AU intensity estimation.

\romannumeral3 ) We evaluate our method\footnote{Code will be released at \url{https://github.com/EvelynFan/FAU}} on two benchmark datasets, showing that our approach performs favorably against related deep models. 

\begin{figure}
\centering
\includegraphics[width=.8\columnwidth]{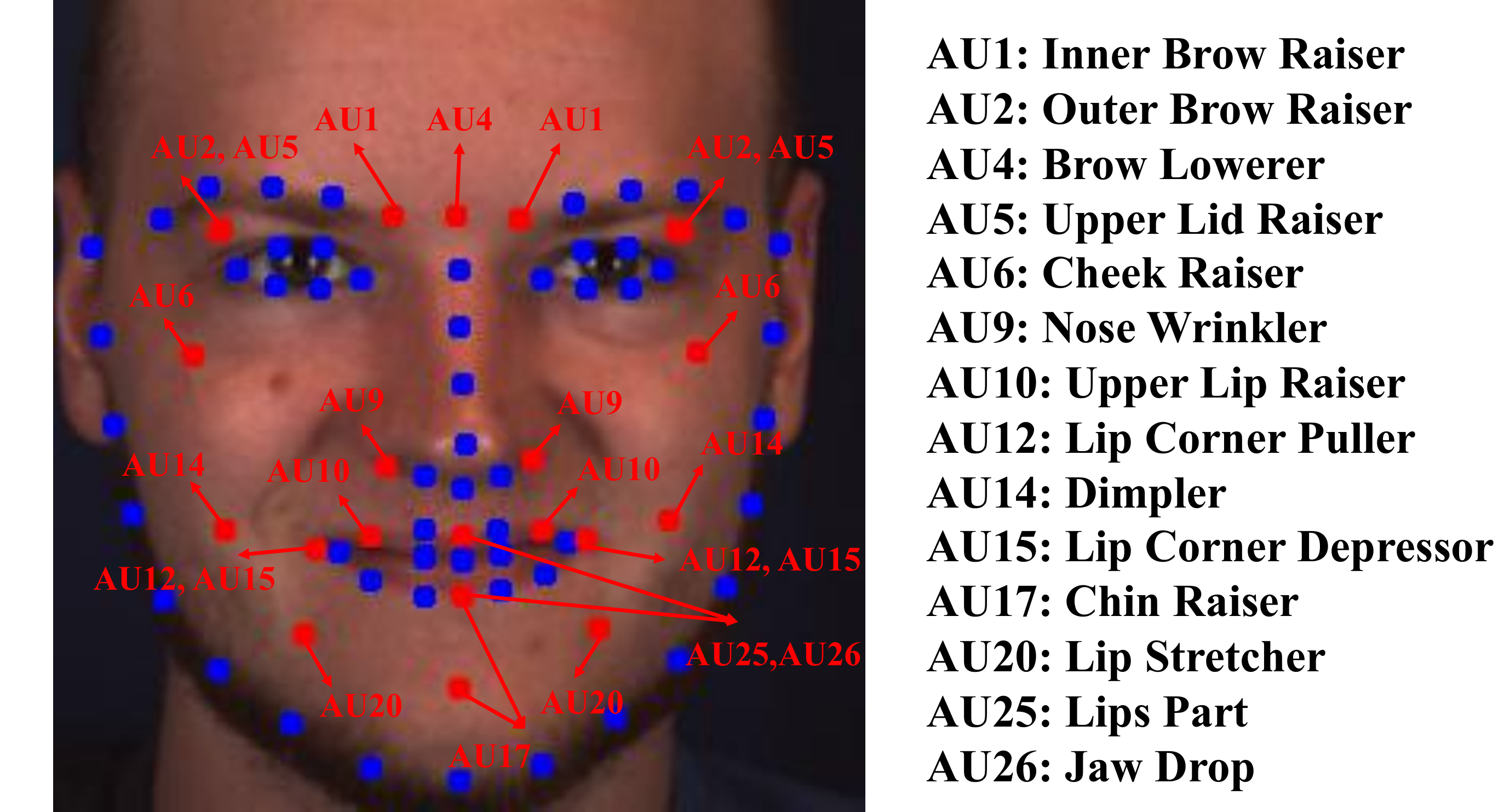}
\caption{\label{fig:location} Central locations of AUs. Different AUs may share the same location, e.g., AU12 and AU15.
}
\end{figure}

\begin{figure*}
\centering
\includegraphics[width=2\columnwidth]{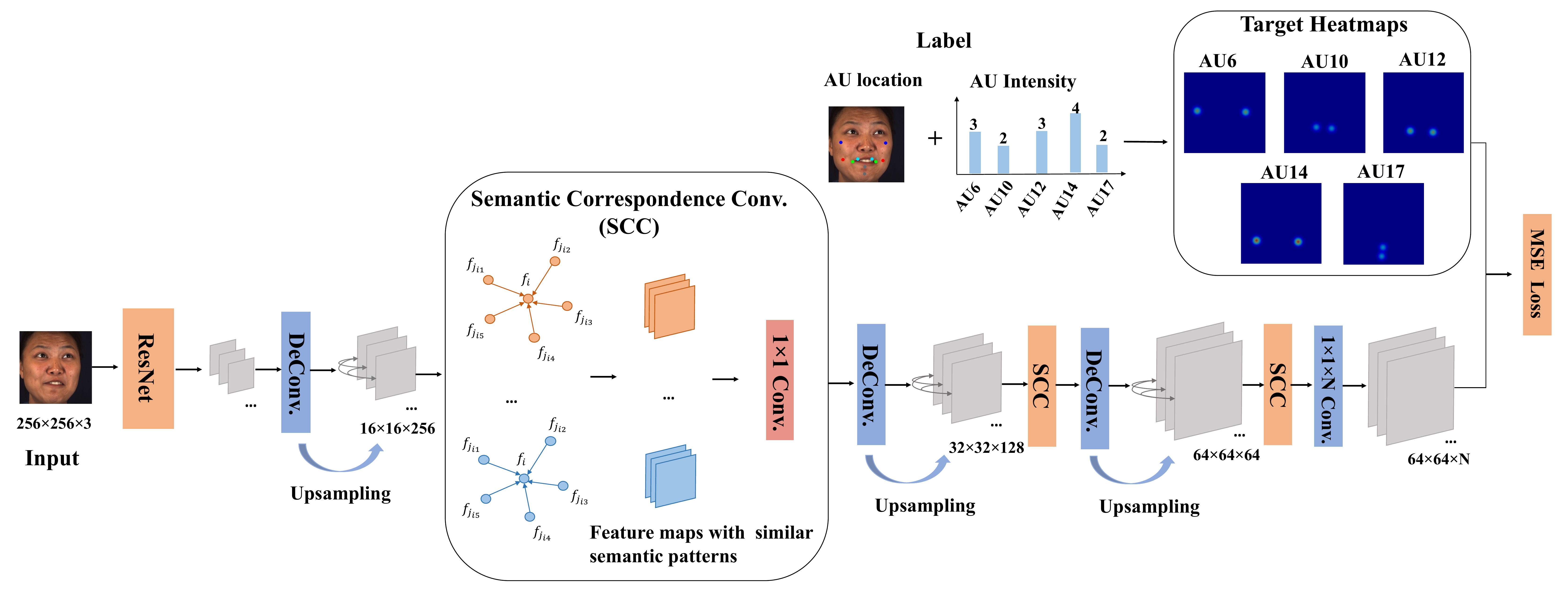}
\caption{\label{fig:framework}Overview of the proposed pipeline for AU intensity estimation. Given the input face image, the outputs are a set of heatmaps for inferring the intensity level of each AU. The ground-truth heatmaps are generated using the Gaussian function determined by AU locations and the corresponding intensities. The semantic correspondence convolutions are integrated to group feature maps sharing similar semantic patterns and to gradually update the relationship graph layer by layer. 
 }
\end{figure*}

\section{Related Work}

To date, most works have focused on facial AU detection. Due to the limited datasets available with AU intensity annotations, few works take into account AU relations for intensity estimation. We review both hand-crafted feature-based works and deep learning-based methods that leverage relationship modeling for AU intensity estimation. 

\subsection{Hand-Crafted Feature-Based Method}

Traditional hand-crafted feature-based AU intensity estimation methods have sought to address this problem based on the probabilistic models, directly capturing the AU dependence. For instance,~\citealp{sandbach2013markov} combined Support Vector Regression (SVR) outputs with AU intensity combination priors in the Markov Random Field (MRF) structures to improve the AU intensity estimation.~\citealp{kaltwang2015latent} formulated a generative latent tree (LT) model, which modeled the higher-order dependencies among the input features and target AU intensities.~\citealp{walecki2016copula} proposed a Copula Ordinal Regression (COR) framework to model the co-occurring AU intensity levels with the power of copula functions and conditional random fields (CRFs). One issue of these shallow model-based methods is that they need to separate feature extraction from model training. To this end, our proposed approach learns the deep representations and semantic relationships jointly in an end-to-end learning system. 

\subsection{Deep Learning-Based Method}

Researchers have just begun to investigate how to leverage deep models that consider AU relations for intensity estimation.~\citealp{walecki2017deep} placed a CRF graph on top of the output layer of a CNN, in order to learn the relations between different AUs using pairwise connections.~\citealp{zhang2018bilateral} exploited relationships among instances and incorporated domain knowledge to learn a frame-level intensity regressor. More recently, the work of~\citealp{wang2018facial} recognized AUs and estimated their intensities via hybrid Bayesian Networks (BNs), where the global dependencies among AUs were captured. Although these deep model-based works can learn better representations, some of them are highly dependent on the probabilistic modeling of AU dependencies. In contrast, our work investigates an alternative method to capture AU dependencies in a heatmap regression framework. With the proposed SCC module incorporated, more complete and high-order AU dependencies are explored.

The model introduced in~\citealp{li2019semantic} learned the semantic relationships with GNN for detecting the presence/absence of AUs. Unlike their method, we consider semantic similarity relationships among feature channels rather than regions. The closest work~\citealp{sanchez2018joint} directly regressed AU locations and intensities through a single Hourglass~\citealp{newell2016stacked}. In our work, our combine heatmap regression and semantic relationships learning in a unified end-to-end framework.

\section{Proposed Methodology}\label{sec:method}

The basic framework is implemented by adding several deconvolutional layers on ResNet~\citealp{xiao2018simple}, as Figure~\ref{fig:framework} shows. Furthermore, the SCC modules are incorporated for dynamically computing the correspondences from multi-resolution feature maps layer by layer. In particular, we predefine the central location of each AU based on the coordinates of facial landmarks in Figure~\ref{fig:location}.

\subsection{Heatmap Regression}

We integrate both the spatial and intensity information of AUs into an end-to-end trained heatmap regression framework. Given a set of images, we formalize the problem as that of jointly predicting multiple AU heatmaps, which potentially encode the intensity of each AU located at a certain position.  As illustrated in Figure~\ref{fig:framework}, each deconvolutional layer is followed with an SCC module that models the relationship among multiple feature maps at this specific resolution level. Finally, the last layer generates a set of $N$ heatmaps for all AUs, where $N$ is the total number of predefined AU locations. The ground-truth possibility heatmap $g_{i}(x)$ for a predefined AU location $L_{i}$ ($i=\{1,\ldots,N\}$) is generated by applying a Gaussian function centered on its corresponding coordinate $\hat{x}_{i}$

\begin{equation}\label{eq-edge}
\begin{split}
g_{i}(x) = \frac{I}{2\pi \sigma ^{2}}\exp (-\frac{\left \| x-\hat{x}_{i} \right \|_{2}^{2}}{2\sigma^{2}}),
\end{split}
\end{equation}
where $I$ is the labeled intensity of the specific AU, and $\sigma$ is the standard deviation. Thus, the probability for $\hat{x}_{i}$ is the largest value in the generated heatmap. If the pixel is farther away from $\hat{x}_{i}$, its probability value would smoothly decrease. In our case, we expect the centre coordinate $\hat{x}_{i}$ can reflect where the specific AU causes appearance changes, whereas $I$ can capture different degrees of the changes.

Let $h_{i}(x; w, b)$ denote the predicted heatmap parametrized by the network weights $w$ and biases $b$. We utilize the $L_{2}$ distance to minimize the difference between $h_{i}(x; w, b)$ and $g_{i}(x)$

\begin{equation}\label{eq-mse}
\begin{split}
L_{MSE} = \min_{w,b} \sum_{i=1}^{N}\sum_{x}\left \| h_{i}(x; w, b) - g_{i}(x) \right \|_{2}^{2}.
\end{split}
\end{equation}

For the inference stage, the highest value of $h_{i}(x; w, b)$ is taken as the predicted AU intensity\footnote{We take the average of those highest values if the AU location is not unique.}, along with its
corresponding location given by $\hat{x}_{i}=\arg \max h_{i}(x; w, b)$.

\begin{figure}
\centering
\includegraphics[width=1\columnwidth]{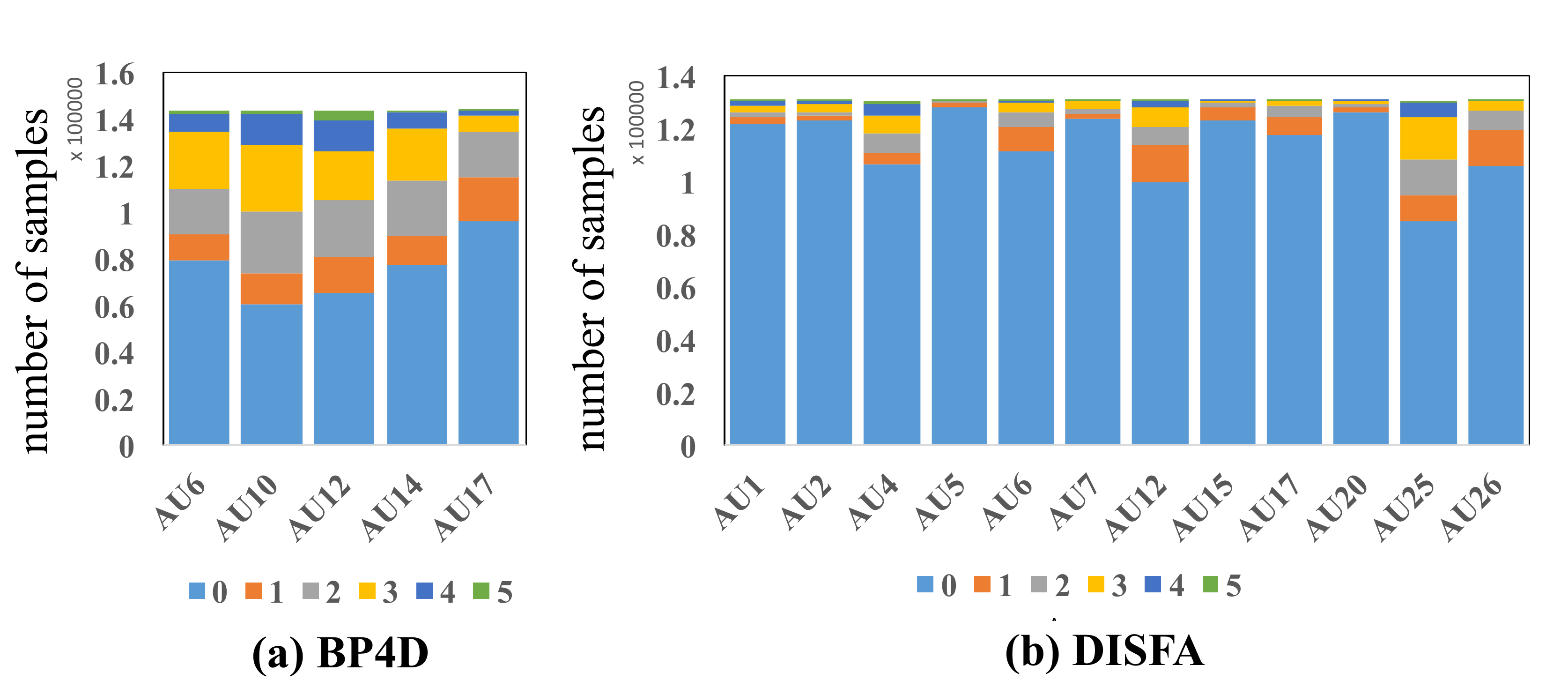}
\caption{\label{fig:distribution} Distribution of the AU intensity levels in the used datasets: BP4D (left) and DISFA (right).
}
\end{figure}

\subsection{SCC: Semantic Correspondence Convolution}

Given the co-occurrences of different AU intensities, the semantic representations of feature maps are highly correlated in spatial distributions. We introduce the semantic correspondence convolution (SCC) operation, aiming to model the correlation among feature channels, where each channel encodes a specific visual pattern of AU. The idea behind the SCC layer is rather simple, mainly inspired by the dynamic graph convolutions in geometry modeling~\citealp{wang2018dynamic}. Intuitively, the feature channels with similar semantic patterns would be activated simultaneously when a specific co-occurrence pattern of AU intensities emerges. In the SCC module, we first construct the k-NN graph by grouping sets of closest feature maps, thus learning to find different co-occurrence patterns. To further exploit the edge information of the graph, we then apply the convolution operations on the edges that connect feature maps sharing similar semantic patterns. Afterwards, the aggregation function, i.e., MAX, is applied to summarize the most discriminative features for improving AU intensity estimation.

\subsubsection{Graph Construction}
From a space with a distance metric, the neighboring feature maps with similar patterns form a meaningful subset. Consider a directed graph $\mathcal G = (\mathcal V, \mathcal E)$, where $\mathcal V = \{1,..., n\}$ and $\mathcal E \subseteq \mathcal V\times \mathcal V$ are the vertices and edges, respectively, the feature maps set is denoted by $F = \{f_{1},f_{2},...,f_{n}\} \subseteq \mathbb{R}$ and the size of each feature map (channel) is given by $M\times M$. In our approach, we rearrange the $M\times M$ feature map in a feature vector with the length of $L = M\times M$. Note that this permutation does not change the feature distribution. $\mathcal G$ is constructed as the k-nearest neighbor (k-NN) graph of $F$, and each node represents a specific feature map. To make use of the edge information, the edge feature is defined by $e_{ij}=h_{\Theta }(f_{i},f_{j})$, where $h_{\Theta }: \mathbb{R}^{L}\times \mathbb{R}^{L}\rightarrow  \mathbb{R}^{{L}'}$ is a nonlinear function with trainable parameters $\Theta$. From the feature space, we regard $f_{i}$ as a central feature and $\{f_{j}:(i,j)\in \mathcal E \}$ as its neighborhood features. We combine the global information encoded by $f_{i}$, with its local neighborhood characteristics, captured by $f_{j}-f_{i}$. Thus, the edge feature function is formulated as 

\begin{equation}\label{eq-edge}
\begin{split}
{e}'_{ijk}=ReLU(\phi_{k} \cdot f_{i}+\omega_{k}\cdot(f_{j}-f_{i})),
\end{split}
\end{equation}
where $(\phi _{1},...,\phi _{K},\omega_{1},...,\omega _{K})$ ($K$ is the number of filters) parameterize $\Theta$,  $\cdot$ represents the inner product, and the ReLU function is selected as $h_{\Theta}$. This function can be implemented with the convolution operator. For each $f_{i}$, the k-NN graph is built by computing a pairwise distance matrix and then taking the closest $k$ feature maps. The pairwise distance matrix is calculated based on the Euclidean distance, which is typically used for measuring the semantic similarity. Besides, we adopt a channel-wise aggregation function, i.e., MAX, to summarize the edge features, as it can capture the most salient features. The output of the SCC module at the i-th vertex is then produced by

\begin{equation}\label{eq-edge}
\begin{split}
{f}'_{ik}=\max _{j:(i,j)\in\mathcal E} {e}'_{ijk}.
\end{split}
\end{equation}

\begin{table*}
\scriptsize
\centering
 \caption{\label{tb:compare} Performance comparison of the models on intensity estimation of 5 AUs from the BP4D dataset, and 12 AUs from the DISFA dataset. $\ast$ indicates results adopted from the corresponding papers for reference. The best results are shown in bold. } 

 \begin{tabular}{ccp{0.25cm}p{0.25cm}p{0.25cm}p{0.25cm}p{0.25cm}p{0.25cm}p{0.25cm}p{0.25cm}p{0.25cm}p{0.25cm}p{0.25cm}p{0.25cm}p{0.25cm}p{0.25cm}p{0.25cm}p{0.25cm}p{0.25cm}p{0.25cm}p{0.25cm}}
 \toprule
  \multicolumn{2}{c|}{ Database}  &\multicolumn{6}{|c|}{ BP4D} &\multicolumn{13}{|c}{ DISFA} \\
  \hline
\multicolumn{2}{c|}{AU}& 6& 10& 12& 14& 17& \multicolumn{1}{c|}{Avg.}& 1& 2 &4 &5 &6 &9& 12& 15& 17& 20& 25& 26& Avg. \\
 \midrule
  \multirow{9}*{ICC}& \multicolumn{1}{c|}{$\text{BORMIR}^{\ast}$ }&  .73 &.68 &.86 &.37 &.47 &\multicolumn{1}{c|}{.62} &.20 &.25 &.30 &.17 &.39 &.18 &.58 &.16 &.23 &.09 &.71 &.15 &.28\\
  & \multicolumn{1}{c|}{$\text{CCNN-IT}^{\ast}$} &.75& .69 &.86& .40& .45& \multicolumn{1}{c|}{.63}&.20& .12& .46 &.08& .48 &.44 &.73& {\textbf{.29}} &{\textbf{.45}}& {\textbf{.21}} &.60 &.46 &.38\\
  & \multicolumn{1}{c|}{$\text{KJRE}^{\ast}$}& .71& .61& {\textbf{.87}}& .39& .42& \multicolumn{1}{c|}{.60}& .27& .35& .25& {\textbf{.33}}& {\textbf{.51}}& .31& .67& .14& .17& .20& .74& .25& .35\\
&\multicolumn{1}{c|}{$\text{KBSS}^{\ast}$}& {\textbf{.76}}& .75& .85& .49& .51& \multicolumn{1}{c|}{.67}& .23& .11& .48& .25& .50& .25& .71& .22& .25& .06& .83& .41& .36\\
&\multicolumn{1}{c|}{ResNet-Deconv} & .70&  .77&  .78& .59  &  .49 &\multicolumn{1}{c|}{.67} & .46 &  .16 & .74& .02& .32& .38&  .71&  .02& .35& .02&  .93& .74 & .40 \\
&\multicolumn{1}{c|}{Hourglass} &  .63 & .70 & .74 &.52 &.38 &\multicolumn{1}{c|}{.59}  & .06 & .04 & .70 & .01 & .28& .43 & {\textbf{.81}}& .02& .26&  .01&  .91& .69 & .35  \\
&\multicolumn{1}{c|}{SCC-Heatmap (Ours)} &.74&  {\textbf{.82}}&  .86& {\textbf{.68}} & {\textbf{.51}}& \multicolumn{1}{c|}{{\textbf{.72}}}& {\textbf{.73}} &  {\textbf{.44}}&{\textbf{.74}} &.06&.27& {\textbf{.51}}&  .71& .04& .37& .04& {\textbf{.94}} & {\textbf{.78}} & {\textbf{.47}} \\
\midrule 
\multirow{9}*{MAE} & \multicolumn{1}{c|}{$\text{BORMIR}^{\ast}$} &.85 &.90 &.68 &1.05 &.79 &\multicolumn{1}{c|}{.85} &.88 &.78 &1.24 &.59 &.77 &.78 &.76 &.56 &.72 &.63 &.90 &.88 &.79\\
& \multicolumn{1}{c|}{$\text{CCNN-IT}^{\ast}$}& 1.17& 1.43 &.97& 1.65&1.08&\multicolumn{1}{c|}{1.26} &.73& .72&1.03& .21& .72& .51& .72& .43& .50& .44& 1.16& .79& .66\\
& \multicolumn{1}{c|}{$\text{KJRE}^{\ast}$}&.82 &.95& .64& 1.08 &.85&\multicolumn{1}{c|}{.87} & 1.02& .92& 1.86& .70& .79& .87& .77 &.60& .80& .72& .96& .94& .91\\
&\multicolumn{1}{c|}{$\text{KBSS}^{\ast}$}&{\textbf{.56}} &.65 &{\textbf{.48}} &.98 &.63 &\multicolumn{1}{c|}{.66} &.48& .49 &.57 &.08 &.26 &.22 &.33 &.15 &.44 &.22 &.43 &.36 &.33\\
&\multicolumn{1}{c|}{ResNet-Deconv} & .61 & .60&.55 &.83 &{\textbf{.39}} & \multicolumn{1}{c|}{.60} & .20 & .16 &.29 & .03 & .29& .16&.33& .15& .20& .08& .31& .36&.21  \\
&\multicolumn{1}{c|}{Hourglass} & .63 &  .70 &.65 & .91& .51 & \multicolumn{1}{c|}{.68}  &  .28 & .26 &  .30 & .03 & .27 & .16&{\textbf{.28}} & .15& {\textbf{.18}} &{\textbf{.08}} & .35& .34&.22  \\
&\multicolumn{1}{c|}{SCC-Heatmap (Ours)} & .61 & {\textbf{.56}} &.52 &{\textbf{.73}} &.50 & \multicolumn{1}{c|}{{\textbf{.58}}} & {\textbf{.16}} & {\textbf{.16}} &{\textbf{.27}} & {\textbf{.03}} & {\textbf{.25}} & {\textbf{.13}} &  .32& {\textbf{.15}} & .20& .09&  {\textbf{.30}}& {\textbf{.32}} & {\textbf{.20}}\\
\bottomrule
 \end{tabular} 
\end{table*}

\subsubsection{Dynamic Graph Update}
In our approach, the dynamic graph convolutions are performed on both low and high resolution feature maps, aiming to capture the high-order AU interactions. Specifically, at the $l$-th layer, a different graph $\mathcal G^{(l)} = (\mathcal V^{(l)}, \mathcal E^{(l)})$ is constructed by computing the closest $k$ feature maps for each single $f_{i}^{(l)}$. The adjacency matrix $\mathcal A^{(l)}$ for $\mathcal G^{(l)}$ is a pairwise distance matrix in the feature space, which represents the relationship among feature maps. Each row stands for a feature map and each entry in $\mathcal A^{(l)}$ indicates the relationship between two corresponding features. The SCC module is flexible since it can take features of arbitrary size as input. It is necessary to constantly keep the graph information updated when new combinations of AU intensity levels happen. Therefore, rather than using a fixed constant graph, our model learns how to construct the graph in each layer during the inference stage.

Overall, the SCC module can be integrated into multiple convolutional layers, and learn to semantically group similar feature channels that would be activated together for a specific co-occurrence pattern of AU intensities.

\subsection{Correspondence with AU Heatmaps}
Generally, our model benefits from the relationship modeling ability of the SCC module, as well as the spatial representation capability of heatmaps. Since AUs can be combined in different intensities to create a vast array of co-occurrences, we would like to explore what the specific visual pattern each channel encodes. As Figure~\ref{fig:framework} shows, following the last SCC module is a $1\times 1$ convolutional layer, where the shape of each learned filter would be $1\times 1\times C \times N$ ($C$ is the number of feature channels and $N$ is the number of output maps). Thus, for each feature channel, there are $N$ weights corresponding to $N$ output maps. Suppose that the number of SCC modules is $L$, the feature maps set generated from the last SCC layer is given by $F^{L} = \{f_{1}^{L},f_{2}^{L},...,f_{C}^{L}\}$. Let $W^{L}_{i} = \{w_{1i}^{L},w_{2i}^{L},...,w_{Ci}^{L}\}$ ($i=1,2,\ldots,N$) denote the $1\times 1$ filter bank for a specific AU, the predicted heatmap $h_{i}$ for this AU is then computed as 
\begin{equation}\label{eq-final}
\begin{split}
h_{i} = F^{L} \otimes W^{L}_{i},
\end{split}
\end{equation}
where $\otimes$ means the tensor product. Hence, $F^{L}$ is better able to explain the AU co-occurrence patterns via the directed $1\times 1$ convolution, capturing the global and evident relations among AUs. Each $f_{j}^{L} \in F^{L}$ ($j=1,2,\ldots,C$) is expected to represent a particular visual pattern, and the corresponding weight $w_{ji}^{L}$ measures its association with $h_{i}$. The weight indicates the probability of the pattern being activated, and a larger one leads to higher intensities of AUs in the pattern. In this way, the final AU co-occurring pattern can be reflected by activating a set of different visual patterns. The additional visualization analysis in the following experiments have provided a more intuitive understanding of the visual patterns.

\section{Experiments}

\subsection{Datasets}
In this study, we conducted experiments on the BP4D~\citealp{zhang2014bp4d} and DISFA~\citealp{mavadati2013disfa} datasets, which provide per-frame AU intensity annotations and have been widely used for AU intensity estimation tasks. BP4D contains 328 facial videos from 41 subjects who were involved in 8 emotion-related elicitation experiments. It provides intensity labels for five AUs: AU6, AU10, AU12, AU14, and AU17, which are quantified into six discrete levels from 0 (absence) to 5 (maximum). In our experiments, we adopted the official training/development sets for evaluation. The DISFA database consists of 27 videos, where each frame is coded with the AU intensity on a six-point discrete scale. There are 12 AUs (1, 2, 4, 5, 6, 9, 12, 15, 17, 20, 25, and 26) annotated by the expert FACS coder. We evaluated the model using the 3-fold subject independent cross-validation protocol, i.e., 18 subjects for training and 9 subjects for testing. The distributions of the intensity levels are reflected in Figure~\ref{fig:distribution}. It is worth noting that they are extremely imbalanced in the six levels.

\subsection{Implementation Details}
 The training and testing processes were performed using NVIDIA GeForce GTX 1080Ti 11G GPUs based on the open-source Tensorflow~\citealp{abadi2016tensorflow} framework. The backbone network was initialized from ResNet-50~\citealp{xiao2018simple}. For the k-NN graph implemented in the SCC layer, we set $k=5$ for the BP4D database and $k=7$ for the DISFA database, which were determined by grid searching. The dlib\footnote{dlib.net} library was utilized to locate the 68 facial landmarks for defining AU locations. The face images were then cropped and resized to $256\times 256$ pixels. During the training phase, we employed the Adam optimizer~\citealp{kingma2014adam}, with an initial learning rate of 5e-4 and a batch size of 32. 

\subsection{Overall Performance}
\subsubsection{Evaluation metrics.} To compare the performance with the state-of-the-art approaches, we use Intra-class Correlation~\citealp{shrout1979intraclass} (ICC(3,1)) and Mean Absolute Error (MAE) for evaluations. Statistically, ICC measures the agreement between $k$ judges who rate $n$ targets of data. In our case, $k = 2$ means that one judge represents AU intensity labels and the other one stands for predicted values, and $n$ denotes the total number of testing examples. We also report MAE, which is widely used for measuring the regression performance.

\subsubsection{Comparison with the State of the Art.}
We compared our model, referred to as SCC-Heatmap, to deep learning-based methods that leverage relationship modeling (CCNN-IT~\citealp{walecki2017deep}, KBSS~\citealp{zhang2018weakly}, BORMIR~\citealp{zhang2018bilateral}) and the recent work KJRE~\citealp{zhang2019joint}. 

Additionally, since our method is based on heatmap regression, we also directly applied two state-of-the-art regression models, Hourglass~\citealp{newell2016stacked} and ResNet-Deconv~\citealp{xiao2018simple}, for AU intensity estimation. Table~\ref{tb:compare} shows the comparative results for the above mentioned methods evaluated on the BP4D and DISFA datasets. The heatmap regression-based methods were evaluated in the same settings by using the source code provided by the authors. The results of other methods are adapted from their corresponding papers. From Table~\ref{tb:compare}, we observe that the proposed SCC-Heatmap performs better with higher ICC and lower MAE in most cases. Specifically, the average ICC of our baseline version (ResNet-Deconv) is $8\%$ higher than Hourglass for the BP4D database. This shows the advantages of ResNet-Deconv in heatmap regression for AU intensity estimation. SCC-Heatmap has an improved performance on both the BP4D and DISFA datasets, which suggests that relationship modeling is crucial. Moreover, compared to other approaches including CCNN-IT, KBSS, BORMIR, and KJRE, for the average ICC, the proposed SCC-Heatmap considering channel-wise feature map relationships shows an increase of $5\%\sim12\%$ on the BP4D database, and an increase of $9\%\sim19\%$ on the DISFA database. The better performance further demonstrates the superiority of the dynamic graph convolutions in capturing AU semantic relationships over other related approaches. To the best of our knowledge, the proposed model achieves the best performance with the highest average ICC, as well as the lowest average MAE, for the BP4D database.

\begin{figure}
\centering         
    \includegraphics[width=.95\columnwidth]{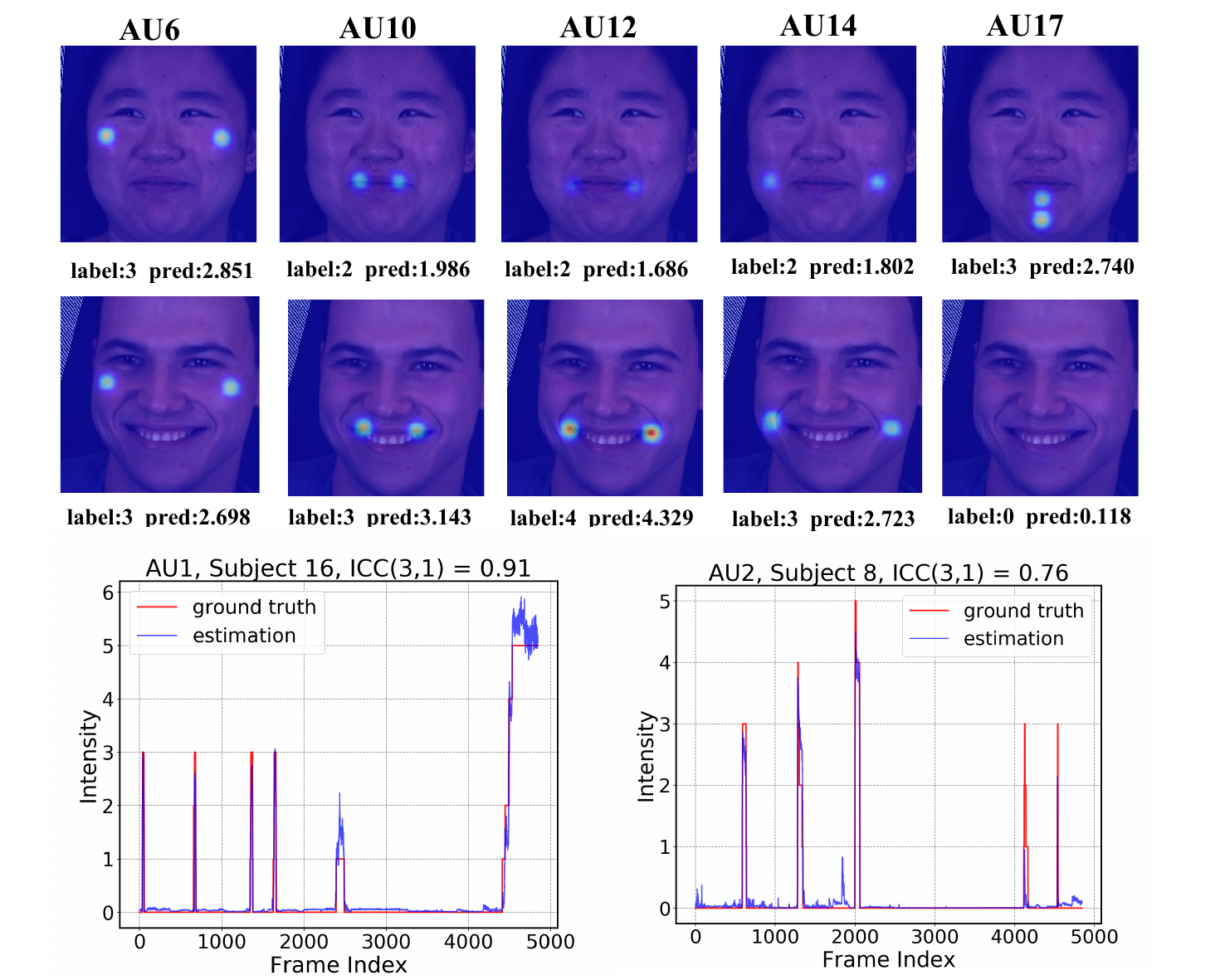}   
\caption{\label{fig:visual}  Two examples of the predicted heatmaps from the BP4D dataset shown in the first two rows. The third row displays two sample results w.r.t. estimated intensities for AU1 and AU2 compared to the corresponding ground-truth from the DISFA dataset. 
    }
\end{figure}

\begin{table}
\centering
 \caption{\label{tb:ablation1} Ablation studies w.r.t. SCC and HR, performed on the testing set of the DISFA dataset. DL = the number of deconvolutional layers. HR = heatmap resolution. SCC = semantic correspondence convolutions.}

 \begin{tabular}{c|ccccc}
  \toprule
 Setting &DL & HR & SCC & ICC & MAE \\ 
 \midrule 
 a& 3 & $64\times 64$ &\cmark& .47 & .20\\
 b& 3 & $64\times 64$ &\xmark & .40 & .21 \\
 c&2 & $32\times 32$ &\cmark& .44 & .23 \\
 d&2 & $32\times 32$ &\xmark& .38 & .25\\
 \bottomrule
 \end{tabular} 
\end{table}

\begin{table}
\centering
 \caption{\label{tb:ablation2} Ablation studies w.r.t. EF, performed on the testing set of the DISFA dataset. EF = edge features.}

 \begin{tabular}{c|ccc}
  \toprule
 Setting &EF  & ICC & MAE \\ 
 \midrule 
 e&  ($f_{i}$, $f_{j}-f_{i}$)& .47 & .20\\
 f& ($f_{i}$,$f_{j}$)  & .43 & .22 \\
 g&  $f_{j}-f_{i}$ & .38 & .21 \\
 \bottomrule
 \end{tabular} 
\end{table}

\subsection{Individual Subject Performance}
We provide the qualitative results in order to further validate the effectiveness of the proposed method. Specifically, given a $256\times 256$ testing image, the network generates a set of $64\times 64$ heatmaps, as illustrated in Figure~\ref{fig:framework}. Two examples of the predicted heatmaps are visualized in Figure~\ref{fig:visual}. We can see that the heatmaps are produced according to the predicted AU locations and their intensities. Then the intensity of each AU is determined by the maximum value of its corresponding heatmap. Following this pipeline, we plot the estimated AU intensities of two subjects from the DISFA dataset which provides per-frame labels, compared to the corresponding ground-truth with titled ICC measure. It can be observed that the estimation line is smooth, stable, and close to the ground-truth. 


\subsection{Ablation Studies}
To verify \textbf{the effectiveness of the semantic correspondence convolutions}, we compared the overall performance of the proposed SCC-Heatmap to the model without considering relationships by removing all the SCC layers. In the meantime, we also investigated \textbf{the effects of the heatmap resolution} by varying the number of deconvolutional layers, which could determine the size of the final output heatmap. Ablation studies were conducted on the DISFA database in four different settings while maintaining all the other settings the same. As shown in Table~\ref{tb:ablation1}, compared to setting (c), setting (a) achieves a higher average ICC by $3\%$ and a lower average MAE by $3\%$. Thus, three deconvolutional layers were used. Settings (b) and (d) show that discarding the SCC module gives a decrease in both ICC and MAE, which suggests that relationship modeling is beneficial in improving the performance of AU intensity estimation. 

We further analyzed \textbf{the effectiveness of the edge features} for constructing the relationship graph. In Equation~\ref{eq-edge}, $f_{i}$ and $f_{j}-f_{i}$ are used to parameterize the edge features, so as to take into account the local characteristics of feature maps while keeping the global information. As illustrated in Table~\ref{tb:ablation2}, setting (e) denotes that both $f_{i}$ and $f_{j}-f_{i}$ are considered, whereas setting (f) and setting (g) discard the local and global information, respectively. As expected, setting (e) leads to a higher ICC value and a lower MAE value, which demonstrates that both the global structure and the local neighborhood information are important for modeling the semantic relationships among feature maps. 

\begin{figure}
\centering 
    \includegraphics[width=.95\columnwidth]{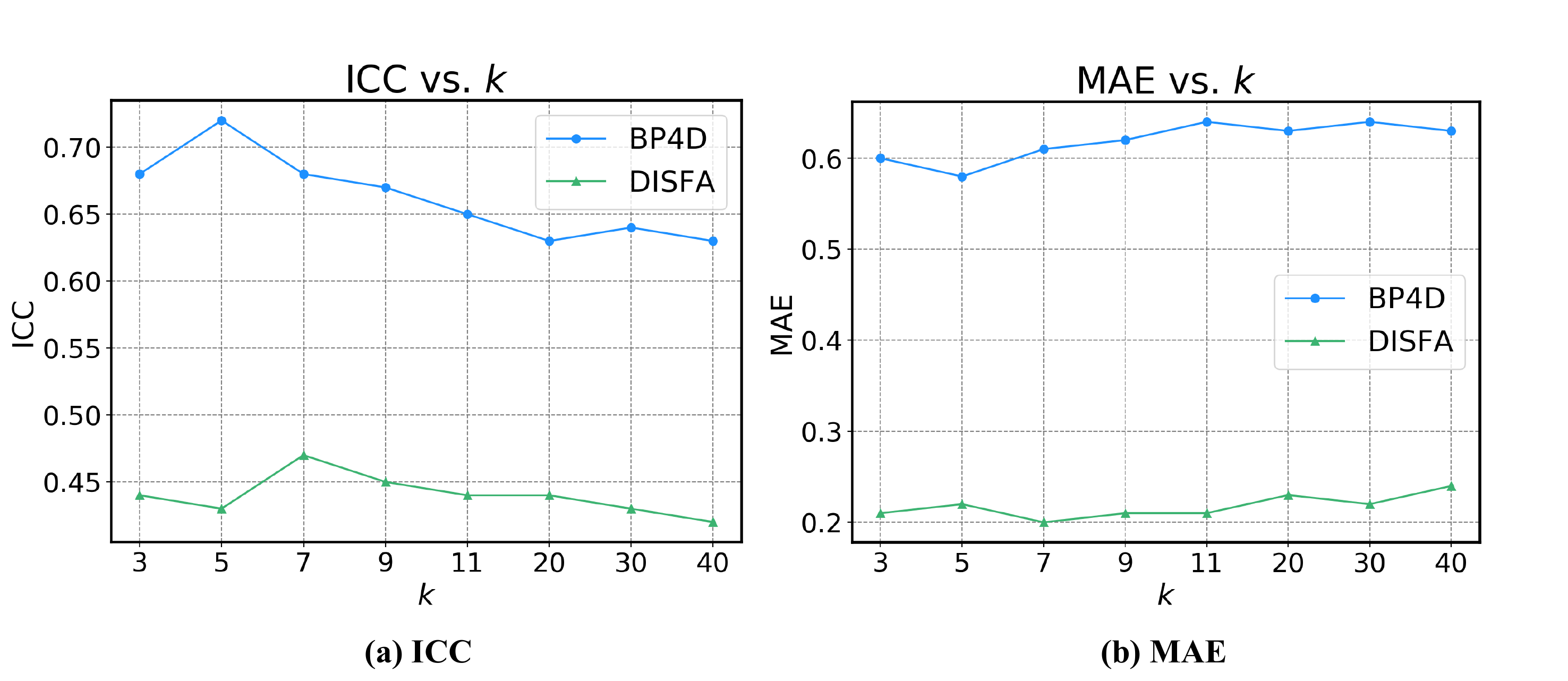}  
\caption{\label{fig:k} Comparison of results with different numbers of nearest neighbors ($k$).}
\end{figure}

To search for an optimal value, we compared the results with different numbers of nearest neighbors $k$, as shown in Figure~\ref{fig:k}. Generally, given that both the ICC and MAE have small fluctuations, the performance is not overly sensitive to $k$. Studying the results, it is notable that larger $k$ gives slightly worse performance. We infer that a larger $k$ might fail to group feature maps appropriately with Euclidean distance.

\subsection{Visualization Analysis}
\begin{figure}
\centering
\includegraphics[width=.95\columnwidth]{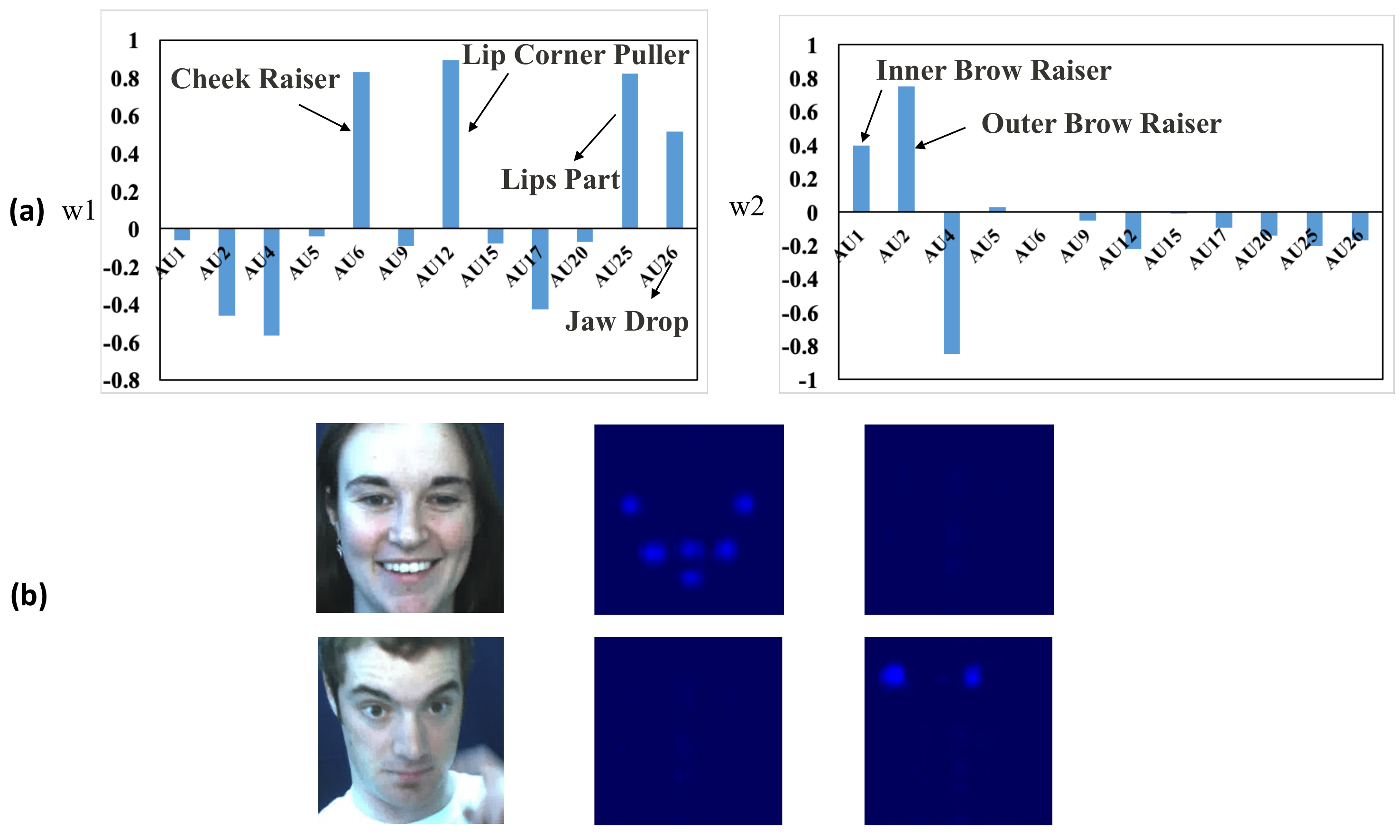}
\caption{\label{fig:relation} (a) Weights $w$ of the first two feature channels in the last SCC layer corresponding to the output maps. (b) From left to right: Input images; Responses of the first feature channel; Responses of the second feature channel.}
\end{figure}

As stated in the previous section, the AU co-occurring pattern is reflected by activating the corresponding feature channels, where each channel encodes a specific visual pattern of AU. To this end, we designed experiments to discover the hidden structure that governs the co-occurrence patterns of AU intensity levels. In Figure~\ref{fig:framework}, through the directed $1\times 1$ convolution operation, the feature maps from the last SCC layer can better represent the patterns of AUs. For the model trained on the DISFA database, the weights $w$ of the first two feature channels in the last SCC layer are visualized in Figure~\ref{fig:relation}. The first feature channel displays the pattern that the face presents ``Cheek raiser'', ``Lip corner puller'', ``Lips part'' and ``Jaw drop'' simultaneously, providing an indication of a positive emotion, e.g., happiness. The second feature channel reflects the pattern that ``Inner brow raiser'' and ``Outer brow raiser'' are likely to occur simultaneously, which suggests the emotions like angry, surprise, etc. Then, we visualize the responses of the first two feature channels using two samples from the DISFA database. It can be observed that one face image activates the first feature channel, whereas the other one activates the second feature channel. Moreover, a larger $w$ leads to a higher response in the location of the specific AU, which indicates higher probability of occurrence. Therefore, different feature channels represent different visual patterns, some of which might be activated together to form the final AU co-occurring pattern.

\begin{figure}
\centering
\includegraphics[width=.95\columnwidth]{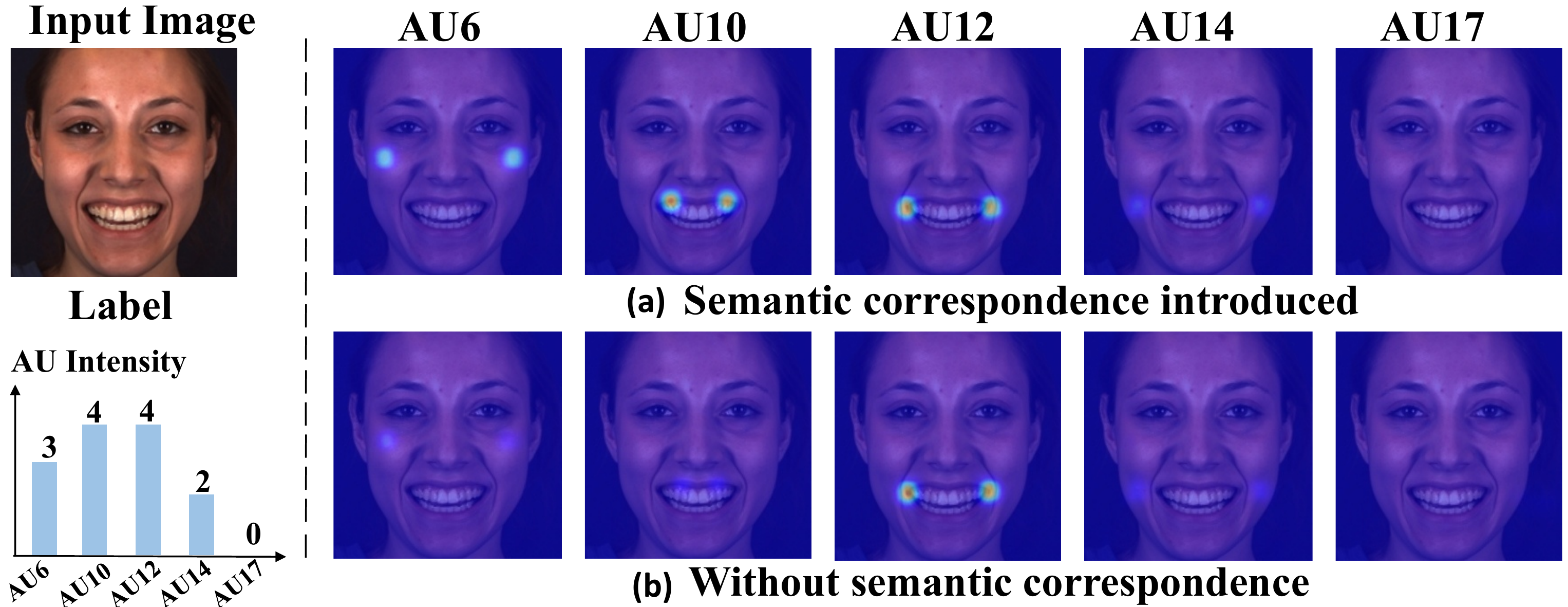}
\caption{\label{fig:compare} Comparison of the predicted heatmaps (a) with, and (b) without semantic correspondence, showing the effectiveness of the semantic correspondence. }
\end{figure}

We also compared the predicted heatmaps between the SCC-Heatmap and the model with all the SCC modules removed in Figure~\ref{fig:compare}. Given an input image from the BP4D dataset, both (a) and (b) predict a high intensity for AU12. While for AU6 and AU10, the refined heatmaps in (a) have much better prediction. We attribute this to the SCC module, where the complementary features are obtained from different channels.
For the model with semantic correspondences considered, we infer that the feature channel associated with AU12 might enhance the response of the feature channel associated with AU6, which validates the effectiveness of the semantic correspondence in modeling the co-occurrence relationships of AU intensities.

\section{Conclusions and Future Work}

In this work, we studied AU intensity estimation from a new perspective by employing the dynamic graph convolutions to capture correlations between neighboring feature maps, thereby modeling the semantic relationships of AU intensity levels. Particularly, the proposed framework keeps updating the relationship graph layer by layer, which can enrich the representation power of AU co-occurrence patterns. In the future, we would like to apply the framework for other tasks, especially those under unsupervised settings. Currently, we choose the k-NN graph to model the relationships. Although it is reasonable for grouping similar feature maps and works well in practice, we plan to investigate alternative approaches in constructing the relationship graph.

\bibliography{6827-AAAI}
\bibliographystyle{aaai}

\end{document}